# TFOC-Net: A Short-time Fourier Transform-based Deep Learning Approach for Enhancing Cross-Subject Motor Imagery Classification

Ahmed G. Habashi[1], Ahmed M. Azab[2], Seif Eldawlatly[1,3*], and Gamal M. Aly[1]

[1]Computer and Systems Engineering Department, Faculty of Engineering, Ain Shams University, Cairo, Egypt

[2]Biomedical Engineering Department, Technical Research Center, Cairo, Egypt

[3]Computer Science and Engineering Department, The American University in Cairo, Cairo, Egypt

*Corresponding author:

Ahmed Habashi

Phone: +21066931570, email: a.habashy@ieee.org.

**Abstract**

Cross-subject motor imagery (CS-MI) classification in brain-computer interfaces (BCIs) is a challenging task due to the significant variability in Electroencephalography (EEG) patterns across different individuals. This variability often results in lower classification accuracy compared to subject-specific models, presenting a major barrier to developing calibration-free BCIs suitable for real-world applications. In this paper, we introduce a novel approach that significantly enhances cross-subject MI classification performance through optimized preprocessing and deep learning techniques. Our approach involves direct classification of Short-Time Fourier Transform (STFT)-transformed EEG data, optimized STFT parameters, and a balanced batching strategy during training of a Convolutional Neural Network (CNN). This approach is uniquely validated across four different datasets, including three widely-used benchmark datasets leading to substantial improvements in cross-subject classification, achieving 67.60% on the BCI Competition IV Dataset 1 (IV-1), 65.96% on Dataset 2A (IV-2A), and 80.22% on Dataset 2B (IV-2B), outperforming state-of-the-art techniques. Additionally, we systematically investigate the classification performance using MI windows ranging from the full 4-second window to 1-second windows. These results establish a new benchmark for generalizable, calibration-free MI classification in addition to contributing a robust open-access dataset to advance research in this domain.

## 1. Introduction

Brain-computer interface (BCI), as one of the most emerging technologies today, could potentially change how humans interact with computers and enhance the quality of life for individuals with disabilities [1]. Among the various BCI paradigms, motor imagery (MI) has gained significant attention due to its non-invasive asynchronous nature and potential applications in rehabilitation and assistive technologies [2]. MI-based BCIs decodes the user's intention to perform a specific motor action, such as imagining moving a hand or a leg, by analyzing EEG signals recorded from the brain non-invasively [3].

Despite considerable advancements in MI-based BCIs, their widespread adoption is hindered by multiple challenges including the inter-subject variability in EEG patterns [4], [5]. This variability necessitates recording subject-specific calibration data to train machine learning classifiers effectively, which can be time-consuming and cumbersome, limiting the usability and accessibility of MI-based BCIs in real-world scenarios. To address this challenge, researchers are actively exploring cross-subject (CS) (i.e. subject-independent) MI-based BCIs [6]. CS classification has emerged as a compelling research direction, aiming to develop models that can generalize across different individuals without requiring subject-specific calibration.

Several studies have directly addressed the challenges of CS classification in MI-based BCIs. For instance, the Separated Channel Convolutional Neural Network (CNN), a novel CNN model, demonstrated the feasibility of classifying MI signals without subject-specific training data, achieving a mean accuracy of 64% on the benchmark BCI Competition IV-2B dataset [7]. Hybrid approaches, such as the Deep Convolutional Recurrent Neural Network, have attempted to capture both spatial and temporal features of EEG signals, but their performance in CS scenarios remains limited, with accuracies not exceeding 54% [8]. The fusion convolutional neural network explored the potential of multi-branch architectures for CS-MI classification, showcasing improved accuracy and robustness on the PhysioNet-eegmmidb dataset [9]. Dolzhikova et al. further explored the application of CNNs for CS-MI classification by proposing an ensemble approach [6]. Their method demonstrated improved performance compared to individual CNN classifiers and several state-of-the-art techniques, achieving accuracies up to 64.16% on the BCI Competition IV-2A dataset and comparable results to the top-performing method on the BCI Competition IV-2B dataset. Roy et al. also tackled the challenge of inter-subject variability in MI-based BCIs using CNNs, introducing the novel concept of 'Mega Blocks' to adapt the network architecture [10]. Their approach focused on continuous decoding of MI signals, crucial for providing real-time neurofeedback in BCI applications. They reported an average inter-subject accuracy of 67.78% on the BCI Competition IV-2b dataset, demonstrating the feasibility of calibration-free MI-BCIs for practical use. The DynamicNet model, while effective for subject-specific classification, experienced a significant drop in performance when applied to CS tasks [11]. Transfer learning and few-shot learning strategies have also been investigated to bridge the gap between source and target subjects, with promising results reported in terms of accuracy and adaptability [12], [13]. However, these methods often involve complex training procedures or require a small amount of target subject data, limiting their practicality in calibration-free scenarios. Despite these efforts, the accuracies achieved in CS-MI classification remain suboptimal, underscoring the need for further exploration and innovation in this domain.

To enhance the representation and analysis of EEG signals, the integration of spectrum image generation (SIG) with deep learning has gained traction in the field of MI-BCI [14], [15], [16], [17], [18]. SIG techniques, which commonly use transformation techniques such as the short-time Fourier transform (STFT) or Fast Fourier Transform (FFT), convert EEG signals into time-frequency representations. These representations resemble images, allowing for the use of advanced image processing and analysis tools to enhance feature extraction, classification and enabling the application of powerful image processing and analysis tools [17], [19]. The STFT is particularly well-suited for non-stationary signals like EEG, where the frequencies of interest might appear and disappear or fluctuate in intensity over the course of a recording [19], [20], [21]. In essence, the STFT is a signal processing technique used to analyze how the frequency content of a signal changes over time [22].

In [18], we explored a calibration-free approach to CS-MI classification using EEG spectrum images and deep learning techniques. We employed a VGG-based CNN for classification and Wasserstein Generative Adversarial Networks (WGAN) for data augmentation. This approach capitalized on converting STFT outputs into 32 × 32 images as input for CNNs, demonstrating the effectiveness of established CNN architectures for EEG-based applications. While this image-based representation offered reliable performance, it also motivated exploration into direct STFT usage without converting it to images. Such direct usage could preserve the richer, inherent time-frequency relationships essential for capturing subtle temporal and spectral details. These insights inspire our current approach, which builds on these advantages to enhance classification accuracy in calibration-free CS-MI.

In this paper, we build upon our previous work in CS-MI classification by proposing a novel Time-Frequency Optimized Classification Network (TFOC-Net) that leverages the strengths of SIG-like representation and deep learning while addressing the limitations of existing methods. Specifically, we introduce several key modifications that aim at improving classification performance and generalizability. First, our approach employs direct classification on STFT-transformed data, comparing it with the base model of STFT-images classification. Second, we examine the impact of the overlap parameter of STFT on the performance and whether varying it can capture finer temporal details in the EEG spectrum. Third, we implement a balanced batching strategy during our CNN training to ensure that each batch contains data from all subjects, promoting better generalization. Notably, our approach aligns with the true definition of "Cross-Subject" classification, as we did not use any data from the test subjects during training. Furthermore, we introduce a novel EEG dataset, meticulously recorded, that will be publicly released, to serve as a valuable resource for future research in CS-MI classification. We compare the performance of our previous and new approaches on this recorded dataset, demonstrating significant improvements. Additionally, we systematically investigate the time segment analysis of CS-MI classification by examining the impact of varying MI segment durations on classification performance. Such an investigation could provide valuable insights into the optimal time windows for extracting discriminative features and inform the design of more efficient and responsive BCIs. Importantly, this work is the first and only study to conduct systematic CS-MI classification using four independent datasets, including three well-established benchmark datasets (IV-1, IV-2A, and IV-2B) and a newly recorded dataset. This extensive validation across multiple datasets strengthens the generalizability and applicability of the proposed method. Furthermore, our approach outperforms previously published methods on the majority of the benchmark datasets, achieving state-of-the-art accuracies, thereby setting a new performance standard in CS-MI classification. Ultimately, we provide comprehensive access to our Python codebase and trained models ([Ahmed Habashy Repo](#)). By combining innovative methodological enhancements with open access to our resources, we aim to contribute to the ongoing pursuit of calibration-free MI-BCIs that can seamlessly adapt to new users and unlock the full potential of this transformative technology.

## 2. Methodology

### 2.1. *Time-Frequency Optimized Classification Network (TFOC-Net)*

Our proposed TFOC-Net enhances CS-MI classification using a novel combination of optimized preprocessing techniques and our VGG-based CNN model. The key components of the approach are summarized as follows:

- Preprocessing with STFT: EEG signals undergo STFT to transform them into time-frequency representations. We increase the STFT overlap to capture finer temporal details in the EEG spectra, ensuring more continuity between segments.

- Direct STFT Classification: We classify the raw STFT-transformed data, preserving the rich frequency and time-related information. This avoids the loss of subtle signal dynamics.

- Balanced Batching Strategy: During CNN training, each batch contains data from all subjects, promoting better generalization across individuals.

- Our CNN model: The STFT-transformed data is classified using a VGG-based CNN, designed to capture complex spatial and temporal features.

The overall pipeline for our approach is illustrated in **Figure 1**, which shows the sequence from raw EEG signal collection, preprocessing via STFT, and classification using the CNN model.

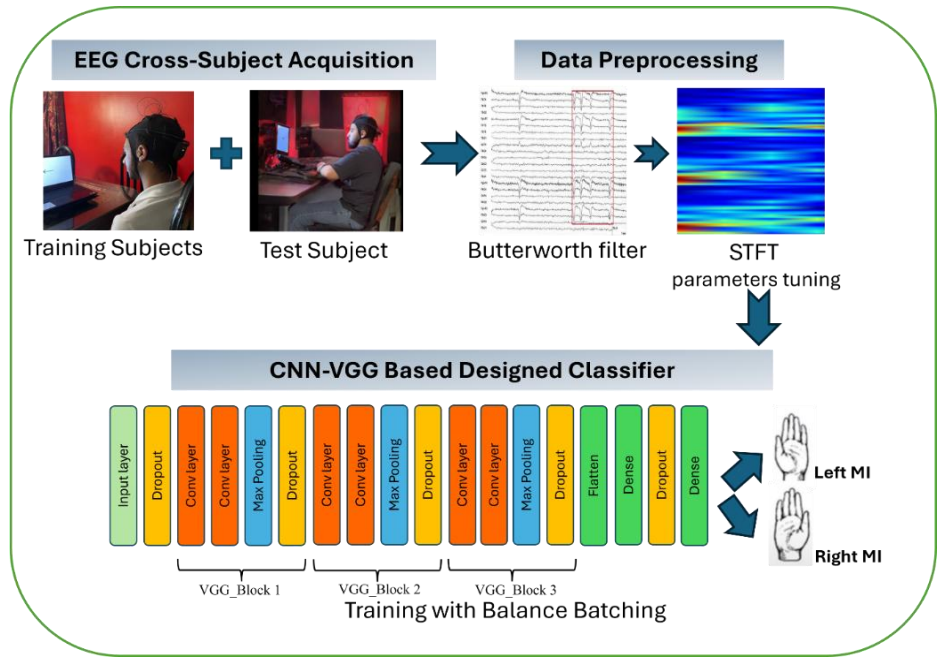

*Figure 1 Schematic of the proposed pipeline for CS-MI classification using VGG-based CNN. The diagram presents the core steps of our approach: (1) Raw EEG Data: Multiple subjects' EEG signals are recorded using an EEG recording system. (2) Preprocessing: The raw EEG signals are filtered using a Butterworth filter and transformed into time-frequency representations using STFT. (3) VGG-based CNN Model: The STFT-transformed data is classified using a VGG-based CNN, designed to capture complex spatial and temporal features. (4) Balanced Batching: Data from all subjects is equally represented during CNN training to improve generalization across subjects.*

*2.2. Benchmark Dataset Description*

This study used three benchmark datasets from BCI Competition IV to validate the proposed framework's effectiveness:

- **BCI Competition IV Dataset 1** [23]: The dataset includes 59 EEG electrodes. We used data from five subjects (subjects b, c, d, g, and e) performing left-hand versus right-hand MI tasks excluding the subject's data representing MI of foot to keep consistency across all used datasets. The data were recorded at a sampling rate of 100 Hz and initially filtered using a 10th-order Chebyshev Type II low-pass filter. We analyzed the data of the three EEG channels (C3, Cz, and C4) only as these channels primarily focus on motor cortex activity [24]. Each subject contributed 100 trials per each class to the dataset.

- **BCI Competition IV Dataset 2A** [25]: This dataset comprises EEG recordings from nine subjects engaged in four distinct MI tasks: left-hand, right-hand, both feet, and tongue movements. For this study, we focus solely on the left-hand and right-hand MI tasks to maintain consistency with the other datasets. The EEG signals were acquired using 22 electrodes placed according to the international 10-20 system, with a sampling rate of 250 Hz. Each subject participated in two sessions on separate days, with each session containing 288 trials (72 trials per class). The data underwent bandpass filtering between 0.5 Hz and 100 Hz. Similar to the other datasets, we focus on the C3, Cz, and C4 channels in our analysis.

- **BCI Competition IV Dataset 2B** [26]: This dataset encompasses EEG recordings from nine participants (B1-B9) engaged in two MI tasks: left-hand and right-hand movements. Data collection involved three bipolar electrodes (C3, Cz, and C4) at a sampling rate of 250 Hz, bandpass filtered in the band 0.5–100 Hz, and a 50 Hz notch filter.

*2.3. Recorded EEG Dataset and Experimental Protocol*

In addition to the benchmark datasets, we recorded a new EEG dataset specifically designed to evaluate our proposed MI classification approach. This dataset was collected using the Unicorn Hybrid Black EEG headset (g.tech medical engineering, Schiedlberg, Austria) [27] with the following setup and experimental protocol.

*2.3.1. Participants*

The dataset consists of EEG recordings from five healthy male volunteers, aged 21 to 25. Participants were chosen based on their lack of neurological disorders and their prior training with motor imagery tasks before recording the dataset to ensure data consistency. All participants were right-handed. The study was conducted in accordance with the Declaration of Helsinki and participants signed an informed consent form before the experiment.

*2.3.2. EEG Recording Setup*

The recorded data included eight EEG channels (Fz, C3, Cz, C4, Pz, PO7, Oz, PO8), along with accelerometer and gyroscope data to capture any head movements or artifacts [1]. However, we focused our analysis only on motor-related cortical areas (C3, Cz, and C4). EEG signals were recorded at a sampling rate of 250 Hz. Real-time monitoring was conducted using the UnicornSuite software to ensure low impedance and optimal signal quality throughout the recording sessions.

*2.3.3. Experimental Protocol*

Participants were instructed to perform MI tasks, specifically imagining the movement of their left hand, and right hand. The protocol, was implemented using the "UnicornSuite_HybridBlack_1.18.00_Win64" software and Python 3.9.7 for GUI, involving the following structure:

- Rest and Preparation: Each trial began with a 2-second rest period, followed by a 2-second cue, with a beeping sound at the beginning lasting for 1.5 sec, indicating which MI task to perform.

- Imagery Period: Participants were then asked to imagine the specified movement for 4 sec.

- Inter-trial Interval: A 1.5-sec rest period was provided between trials to prevent mental fatigue and ensure clear separation of EEG signals across trials. The visual sequence of these trial phases is depicted in **Figure 2a**.

- Number of Trials: each subject participated in two sessions. The number of trials varied across subjects and sessions due to differences in individual time constraints, fatigue levels, and the specific objectives for each participant. Subject 1 completed 40 training and 40 evaluation trials in each session. Subject 2 completed 60 training and 20 evaluation trials in each session. Subject 3 completed 150 training trials in the first session and 40 evaluation trials in the second session. Subjects 4 and 5 each completed 150 training and 150 evaluation trials in each session.

Cues were presented visually on a 15-inch computer screen, with arrows indicating the direction of the imagined movement. The order of MI tasks was randomized within each session to mitigate order effects and ensure the robustness of the recorded data.

### 2.3.4. Dataset Availability

The recorded EEG dataset will be made publicly available, providing the research community with a new resource for studying cross-subject MI classification. This dataset includes the raw and preprocessed EEG data, task labels, and detailed metadata to facilitate reproducibility and further research.

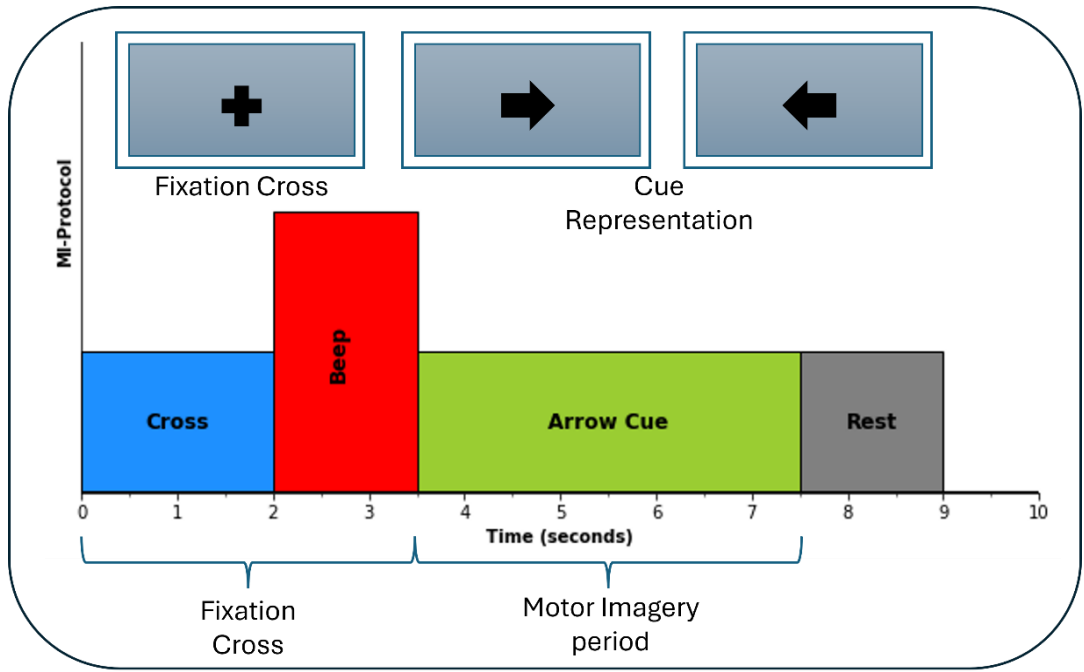

(a)

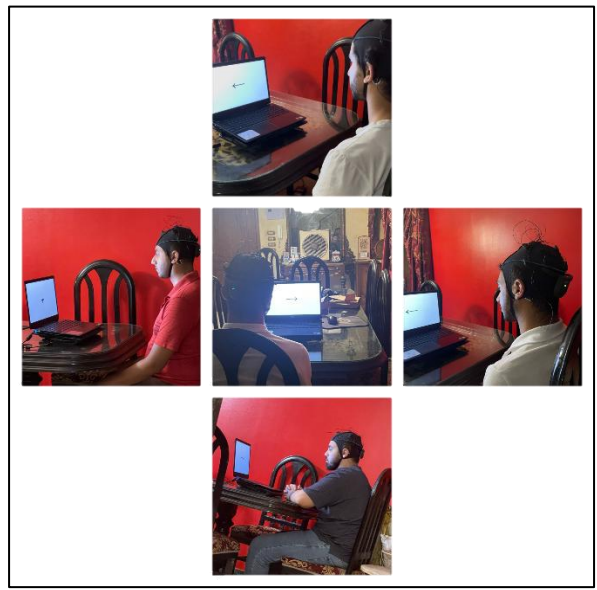

(b)

*Figure 2: (a) Illustration of the motor imagery (MI) trial protocol used in the study, showing the sequence of events from rest and preparation, cue presentation, to imagery period and inter-trial interval. This figure outlines the structure and timing of the tasks performed by participants during EEG recording. (b) The Subjects during recoding MI tasks*

## *2.4. Preprocessing Pipeline*

We used a 6$^{th}$ order Butterworth in the band 8-30 Hz to filter EEG signals [28]. STFT works by dividing a signal into short, overlapping segments (windows) and computing the Fourier Transform of each segment. This produces a series of spectra, each representing the frequency content of the signal at a particular time. The amount of overlap between windows and the window size itself are key parameters that can significantly affect the results. The window size for both datasets (IV-2A, IV-2B and our Recorded Dataset) has been set to 256 samples and 128 for dataset (IV-set 1) [29], [15], [17], [18]. For each trial, we used the raw transformed STFT data to preserve all relevant information compared to the base model that converts STFT outputs into 32 × 32 images as input for CNNs. The STFT results in a complex-valued matrix and its magnitude representing time-frequency components. By directly using this data, the classifier can utilize all available information which is essential for capturing the complex dynamics of EEG signals.

## 2.5. *TFOC-Net Architecture*

The CNN-VGG based model we propose is specifically designed to classify EEG data after it has undergone STFT preprocessing. The Visual Geometry Group (VGG) network architecture, introduced by Simonyan and Zisserman in 2014 [30], marked a significant milestone in the evolution of CNNs. The hallmark of VGG lies in its simplicity and depth, achieved by stacking multiple small 3 × 3 convolutional layers. This design philosophy, a departure from the trend of using larger filters, proved remarkably effective in image classification tasks, demonstrating that depth can be a potent weapon in feature learning [30]. Our model is based on the architecture from our previous work [18], incorporating modifications to effectively accommodate the updated input data shape and capture the features in the STFT-transformed EEG data. The input to the CNN is a 2D array representing the STFT of the EEG data, with a shape of (number of frequency bins, number of time points × number of channels). The frequency bins correspond to the frequency components derived from the STFT, while the time points represent the temporal resolution. The final dimension is for the single channel of the input "image".Table 1 illustrates the parameters of the classifier. The network starts with a dropout layer to reduce overfitting by randomly setting a fraction of the input units to zero during training. This is followed by two convolutional layers, each with 32 filters and a kernel size of 3 × 3, which help the model learn spatial hierarchies of features. We then apply a MaxNorm constraint to the kernel weights in these layers to limit their magnitude, promoting better generalization. After another dropout layer, the network includes two more convolutional layers with 64 filters each, again with a kernel size of 3 × 3. A third dropout layer is applied before the final set of convolutional layers, which have 128 filters each and a kernel size of (3 × 3).

Once the convolutional layers have done their work, the output is flattened into a one-dimensional vector. This vector is then passed through a dense layer with 256 units and Rectified Linear Unit (ReLU) activation, mapping the learned features into a decision space [31]. Another dropout-layer follows to further reduce overfitting. The final output layer consists of 2 units with a softmax activation function, providing class probabilities for binary classification. For training, we use the Root Mean Squared Propagation (RMSprop) optimizer and sparse categorical cross-entropy as the loss function [32], [33]. To avoid overfitting, we employ early stopping and checkpointing strategies, saving the best model based on validation accuracy [34].

*Table 1 Detailed architecture of the CNN model used for EEG classification after STFT preprocessing, including the layer types, filter sizes, number of filters, activation functions, and dropout rates. This architecture is designed to capture spatial and temporal features from EEG data, enhancing CS-MI classification performance.*

| Layers | Filter Size | Number of Filters | Activation | Dropout rate | Other Parameters |
|---|---|---|---|---|---|
| Dropout | - | - | - | 0.2 | input shape = in raw |
| Conv2D | (3, 3) | 32 | ReLU | - | padding='same', kernel initializer='he_uniform', kernel constraint=MaxNorm (3) |
| Conv2D | (3, 3) | 32 | ReLU | - | padding='same', kernel initializer='he_uniform', kernel constraint=MaxNorm (3) |
| Dropout | - | - | - | 0.2 | - |
| Conv2D | (3, 3) | 64 | ReLU | - | padding='same', kernel initializer='he_uniform', kernel constraint=MaxNorm (3) |
| Conv2D | (3, 3) | 64 | ReLU | - | padding='same', kernel initializer='he_uniform', kernel constraint=MaxNorm (3) |
| Dropout | - | - | - | 0.4 | - |
| Conv2D | (3, 3) | 128 | ReLU | - | padding='same', kernel initializer='he_uniform', kernel constraint=MaxNorm (3) |
| Conv2D | (3, 3) | 128 | ReLU | - | padding='same', kernel initializer='he_uniform', kernel constraint=MaxNorm (3) |
| Dropout | - | - | - | 0.4 | - |
| Flatten | - | - | - | - | - |
| Dense | - | 256 | ReLU | - | kernel initializer='he_uniform', kernel constraint=MaxNorm (3) |
| Dropout | - | - | - | 0.4 | - |
| Dense | - | 2 | Softmax | - | kernel initializer='he_uniform' |

We developed the model in TensorFlow, and the experiments were conducted on a machine equipped with Intel Core i7-10750H and 32 GB of RAM. The training was accelerated by a NVIDIA GeForce RTX 2070 graphics processing unit.

### 2.6. Training and Evaluation

For training and evaluation, we adopted the Leave One Subject Out (LOSO) cross-validation approach, which is one of the most rigorous and widely accepted methods for evaluating machine learning models in EEG classification tasks [7], [12], [35]. This approach ensures that the model performance is tested on data from subjects it has never encountered during training, providing a robust measure of generalization. In LOSO, the data from one subject is entirely excluded from the training process and is used exclusively for testing. The model is trained on data from all other subjects, and this process is repeated for each subject in the dataset. The final performance metric is averaged across all these iterations, offering a comprehensive evaluation of the model's ability to generalize to unseen subjects.

Unlike other studies that sometimes incorporate a portion of the test subject's data into the training process, thus introducing potential bias, we strictly avoided using any data from the test subjects during training. TFOC-Net enhances the credibility of our results, as it simulates a real-world scenario where the model must perform on completely novel data. The CNN model was trained for 200 epochs with a batch size of number of training subjects multiplied by 4 through the three benchmark datasets.

To evaluate the performance of the model, we primarily relied on accuracy as a measure of the classifier's effectiveness in distinguishing between different MI classes. Accuracy, a commonly used metric in classification tasks, measures the proportion of correctly classified instances out of the total number of instances. Additionally, to assess the statistical significance of the results, we conducted a Wilcoxon rank-sum test. This non-parametric test was chosen due to its ability to compare differences between two independent samples, making it particularly suitable for evaluating performance differences across subjects in EEG-based classification tasks [36].

## 3. Results

### 3.1. TFOC-Net vs Direct Classification Post STFT

We first examine the impact of direct classification of STFT output compared to the base model which uses image-based classification after applying STFT to EEG data. **Figure 3**, **Figure 4**, and **Figure 5** illustrate the CS-MI classification accuracy results for the three benchmark datasets (IV-1), (IV-2A), and (IV-2B), respectively, during different phases of our approach.

Across all three datasets, a consistent trend emerged: direct classification of STFT data led to improved accuracy compared to the image-based method. The most striking improvement was seen in dataset IV-2B, where the mean accuracy jumped from 67.62% to 78.54%, a substantial increase of nearly 11%. This improvement was statistically significant ($p < 0.01$, Wilcoxon signed-rank test), underscoring the advantage of direct STFT classification.

For dataset IV-1, the mean accuracy increased from 59.2% with images to 60.29% with direct STFT classification. The largest subject enhancement was observed for subject E, whose accuracy increased from 64% to 83%. While this improvement was not statistically significant ($p = 0.69$, Wilcoxon signed-rank test), it still suggests a potential benefit of the direct STFT classification approach. Similarly, for dataset IV-2A, the accuracy increased from 63.19% to 64.81% where subject A03 showed the most significant individual improvement, with accuracy increasing from 65.97% to 73.61%. The relatively smaller improvements in these datasets might indicate that the image-based approach was already capturing a significant portion of the relevant information, leaving less room for gains through direct STFT classification. Nevertheless, the consistent trend across all datasets reinforces the advantage of the direct classification method.

### 3.2. Impact of STFT Parameter Optimization and Balanced Batching

One of the critical modifications in our preprocessing pipeline was adjusting the overlap between consecutive windows in the STFT and implementing a balanced batching strategy during CNN training. The overlap parameter controls the number of overlapping points between consecutive STFT segments, influencing the trade-off between time and frequency resolution. Balanced batching ensures that each training batch contains data from all training subjects, promoting better generalization.

The results in **Figure 3**, **Figure 4**, and **Figure 5** reveal that increasing the STFT overlap generally led to further improvements in classification accuracy. The impact of increased overlap was most evident in dataset IV-1, where the mean accuracy increased from 60.29% to 67%, a statistically significant increase ($p < 0.05$, Wilcoxon signed-rank test). In dataset IV-2A, the increase in overlap led to a smaller but still appreciable gain in mean accuracy, from 64.81% to 65.33%. The most notable individual improvement was for subject A03, whose accuracy increased from 73.61% to 81.25%. For dataset IV-2B, a more modest improvement from 78.54% to 79.2% was observed.

Finally, the implementation of balanced batching consistently yielded the best results across all datasets. In dataset IV-1, it pushed the mean accuracy to 67.6%. Subject D showed the largest improvement, from 63.5% to 65.5%. In dataset IV-2A, balanced batching resulted in a mean accuracy of 65.96%, with subject A01 improving from 65.28% to 67.36%. In dataset IV-2B, the mean accuracy reached 80.22%, and subject B04's accuracy increased further from 92.51% to 94.79%. These results collectively demonstrate the efficacy of the proposed modifications in enhancing the performance of CS-MI classification using our CNN model. The combination of these techniques contributes to the superior performance observed in the final results.

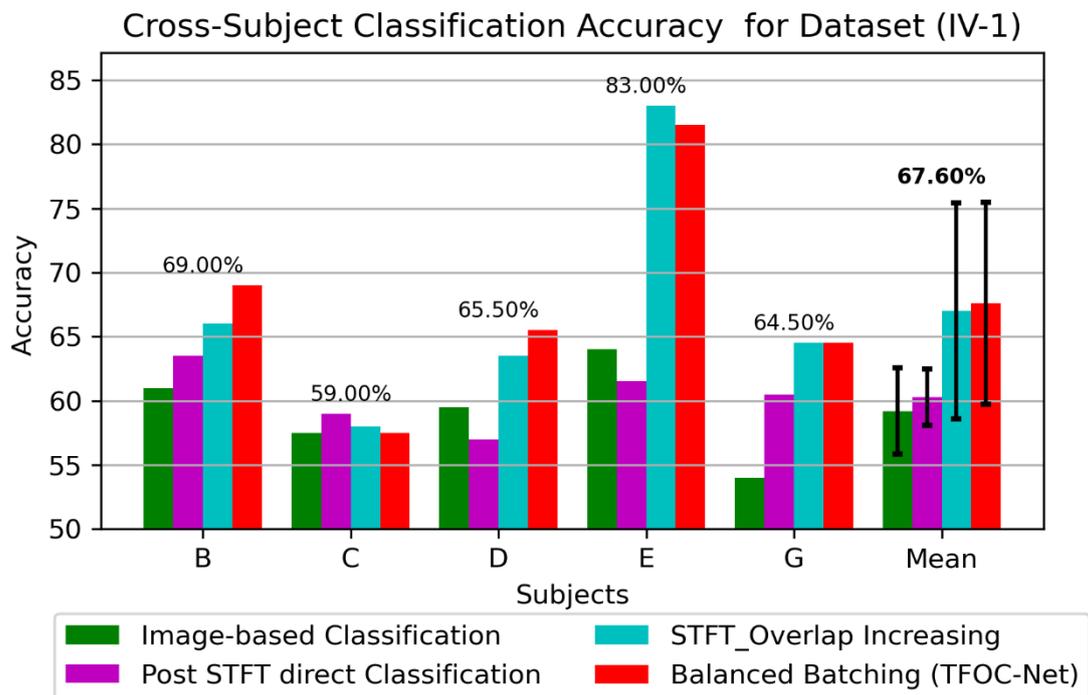

Figure 3 Classification accuracy of CS-MI tasks on the BCI Competition IV Dataset 1 during various modification stages of our approach, including transitions from image-based classification to direct STFT, increased STFT overlap, and balanced batching. The table highlights the impact of these modifications on the performance across different subjects

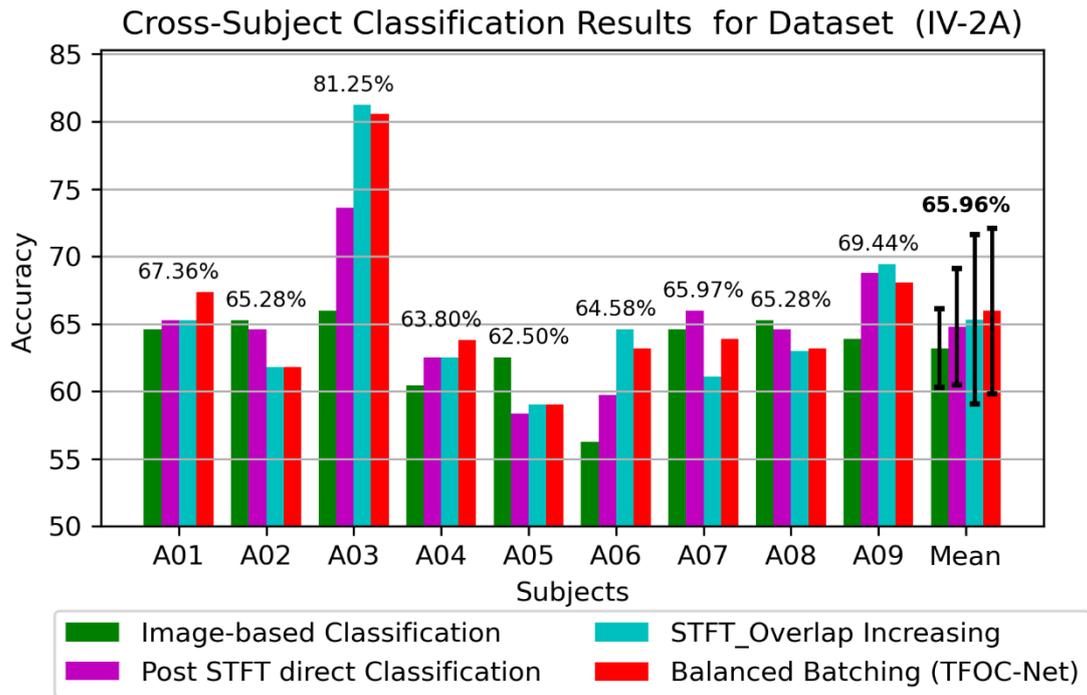

*Figure 4 CS classification accuracies on BCI Competition IV Dataset 2A showing the performance improvements at each modification stage—direct STFT classification, increased overlap in STFT, and balanced batching. This table demonstrates the effectiveness of our approach across various subjects.*

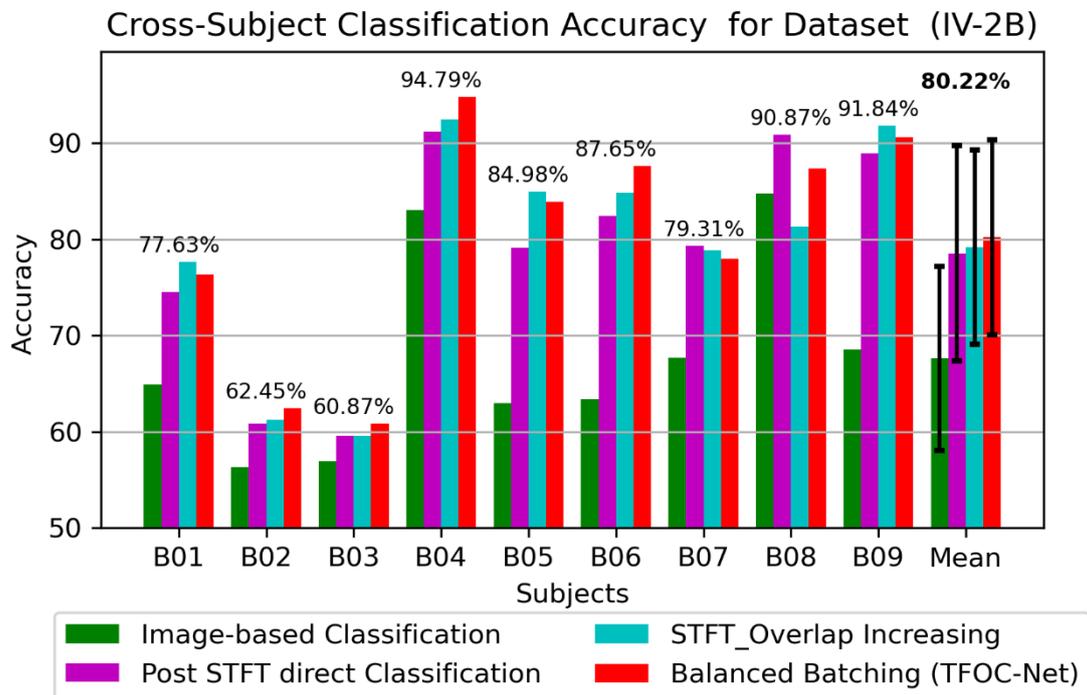

*Figure 5 Performance analysis of CS classification accuracy on BCI Competition IV Dataset 2B through different stages of our methodology. The table shows accuracy improvements with the transition from image-based to direct STFT classification, with further gains from optimized STFT parameters and balanced batching strategies.*

## 3.4. Time Segment Analysis in CS-MI Classification

In this section, our primary objective is to explore usability enhancement by reducing the window size for data collection, which minimizes the burden on BCI users. We analyze the effect of using different time segments within the MI period on CS classification accuracy. Specifically, we investigate the classification performance using segments of 4 seconds (the full MI period), 3 seconds (the first 3 seconds), 2 seconds (the first, middle, or last 2 seconds), and each individual second separately. This analysis was conducted for all three benchmark datasets: BCI Competition IV Dataset 1, Dataset 2A, and Dataset 2B.

For BCI Competition IV Dataset 1, the classification accuracy for different MI periods is shown in **Table 2**. The results demonstrate a significant variation in performance based on the segment of the MI period used. For the full 4-second period, the mean accuracy is 67.60%. Interestingly, the first 3 seconds yield a slightly lower accuracy (63.6%), while the middle 2-second segment results in the highest mean accuracy (64.6%) across the 2-second segments examined. This suggests that critical MI information may be concentrated in the middle of the trial. Additionally, individual seconds within the MI period show varying performance, with the third second producing the highest accuracy (65.09%), indicating that time segment analysis plays a role in the strength of the EEG signal for MI classification.

The classification results for Dataset 2B, shown in **Table 3,** reveal a similar pattern. The highest mean accuracy is achieved when using the full 4-second MI segment. However, using shorter segments, particularly the first 3 seconds or the first 2 seconds, also yields competitive results with mean accuracies of 79.46% and 77.475%, respectively. This suggests that a significant portion of the discriminative information for MI classification is concentrated in the early part of the MI period. As the segment duration decreases or shifts towards the later part of the MI period, the performance generally declines, indicating a potential decrease in the signal-to-noise ratio or a change in the nature of the neural activity associated with MI.

Finally, **Table** 4 displays the classification results for Dataset 2A. The full 4-second window achieves a mean accuracy of 65.96%. Notably, the middle 2-second segment produces a slightly higher accuracy (63.96%) compared to the first or last segments. This suggests that the optimal time for MI classification may vary depending on the subject, but consistent patterns emerge when examining the mid-range time windows.

*Table 2 Classification accuracy for CS-MI tasks on BCI Competition IV Dataset 1, comparing performance across various MI time segments. Results are presented for the full 4-second MI window, the first 3 seconds, and different 2-second and 1-second segments. The analysis highlights the influence of temporal segmentation on classification accuracy across different subjects.*

| Subject | 4-sec | 3-sec (1st) | 2-sec (1st) | 2-sec (mid) | 2-sec (end) | 1st sec | 2nd sec | 3rd sec | 4th sec |
|---|---|---|---|---|---|---|---|---|---|
| B | **69** | 66.5 | 59 | 60.5 | 65 | 60 | 57.5 | 63.5 | 62 |
| C | 57.5 | 55.5 | 61 | 59.5 | 58.5 | 58 | **59.5** | 60.5 | 60 |
| D | **65.5** | 63 | 63.5 | 63 | 63 | 60.5 | 61.5 | 65 | 60.5 |
| E | **81.5** | 79 | 70.5 | 76.5 | 79 | 64 | 70 | 72 | 68 |
| G | **64.5** | 57 | 57.5 | 63.5 | 54 | 60 | 57.5 | 64.5 | 59.5 |
| Mean | **67.60 ±7.89%** | 63.6 ±5.37 | 62.2 ±4.59 | 64.6 ±6.13 | 63.89 ±8.45 | 60.5 ±1.94 | 61.2 ±4.64 | 65.09 ±3.78 | 62 ±3.11 |

*Table 3 CS classification accuracy on BCI Competition IV Dataset 2B across different MI time windows. Performance is compared for the full 4-second window, the first 3 seconds, and specific 2-second and 1-second segments, illustrating the effect of temporal segmentation on classification outcomes.*

| Subject | 4-sec | 3-sec (1st) | 2-sec (1st) | 2-sec (mid) | 2-sec (end) | 1st sec | 2nd sec | 3rd sec | 4th sec |
|---|---|---|---|---|---|---|---|---|---|
| B01 | 76.32 | 77.19 | 71.05 | 74.123 | 71.053 | 73.684 | 61.404 | 68.421 | 65.789 |
| B02 | **62.45** | **64.08** | 62.85 | 57.551 | 55.918 | 63.265 | 56.735 | 57.551 | 56.734 |
| B03 | **60.87** | 59.13 | 60.86 | 58.696 | 58.696 | 59.130 | 57.826 | 56.522 | 60.437 |
| B04 | **94.79** | 93.48 | 93.159 | 85.668 | 78.827 | 94.463 | 86.971 | 76.873 | 69.381 |
| B05 | 83.89 | 82.78 | 77.655 | 78.388 | 71.062 | 75.824 | 74.359 | 70.330 | 58.974 |
| B06 | **87.65** | **88.45** | 82.47 | 75.697 | 76.096 | 78.884 | 70.518 | 74.502 | 66.135 |
| B07 | 78.01 | 77.16 | 75.86 | 75.862 | 67.672 | 74.569 | 72.414 | 65.517 | 60.345 |
| B08 | **87.39** | 83.48 | 84.783 | 80.435 | 73.043 | 82.174 | 86.087 | 74.783 | 70.870 |
| B09 | 90.61 | 89.38 | 88.571 | 84.898 | 84.082 | 83.265 | 80.0 | 79.591 | 75.102 |
| Mean | **80.22 ±10.11** | 79.46 ±10.87 | 77.475 ±10.43 | 74.59 ±9.56 | 70.716 ±8.5 | 76.14 ±9.69 | 71.81 ±10.75 | 69.39 ±7.72 | 64.862 ±5.82 |

*Table 4 CS classification accuracies on BCI Competition IV Dataset 2A, analyzing performance across multiple MI time segments. The table compares accuracy for the full 4-second MI window, the first 3 seconds, and various 2-second and 1-second intervals, providing insights into the time segment analysis of MI classification.*

| Subject | 4-sec | 3-sec (1st) | 2-sec (1st) | 2-sec (mid) | 2-sec (end) | 1st sec | 2nd sec | 3rd sec | 4th sec |
|---|---|---|---|---|---|---|---|---|---|
| A01 | **67.36** | 66.67 | 63.889 | 62.5 | 60.417 | 61.805 | 62.5 | 65.277 | 55.55 |
| A02 | 61.81 | **64.58** | 61.111 | 58.333 | 59.72 | 60.416 | 59.027 | 60.416 | 54.86 |
| A03 | 80.56 | **82.64** | 70.833 | 70.139 | 63.88 | 73.61 | 66.667 | 63.889 | 61.81 |
| A04 | 63.8 | 61.8 | **67.361** | 64.583 | 56.94 | 65.277 | 61.806 | 54.86 | 60.417 |
| A05 | **59.03** | 57.64 | 55.556 | 58.333 | 59.72 | 54.86 | 59.722 | 55.556 | 57.638 |
| A06 | **63.19** | 60.04 | 61.111 | 61.111 | 60.417 | 62.5 | 59.722 | 59.72 | 61.11 |
| A07 | **63.89** | 63.89 | 63.89 | 61.111 | 56.944 | 60.417 | 58.333 | **65.97** | 54.86 |
| A08 | 65.972 | 65.28 | 65.28 | 65.28 | 59.028 | 66.667 | **70.139** | 61.11 | 56.94 |
| A09 | 68.07 | 64.58 | 65.278 | 74.306 | 69.44 | **69.44** | 63.889 | 63.889 | 68.05 |
| Mean | **65.96 ±6.14%** | 65.35 ±6.61 | 63.81 ±4.07 | 63.96 ±5.04 | 60.72 ±3.64 | 63.88 ±5.23 | 62.42 ±3.69 | 61.19 ±3.78 | 59.03 ±4.06 |

## 3.5. Recorded Dataset Classification Accuaracies

The efficacy of the proposed approach was further validated on our newly recorded EEG dataset. The results obtained on this dataset are presented in **Figure 6**, showcasing the classification accuracies achieved by both the base model and TFOC-Net. The direct classification approach after STFT led to an immediate improvement in mean accuracy, rising from 57.37% to 60.81%. This gain highlights the benefits of preserving the raw STFT data, allowing the classifier to access richer time-frequency representations. Increasing the STFT overlap further enhanced performance, boosting the mean accuracy to 62.82%. The most significant individual improvement was observed for Subject 2, whose accuracy increased by 10% with this modification. Finally, the balanced batching strategy led to the best overall performance, achieving a mean accuracy of 63.54%. Subject 3 showed the most notable gain with this strategy, improving from 60% to 61.05%. In summary, the proposed approach consistently outperformed the previous method across all subjects, demonstrating a clear improvement in classification accuracy. The mean accuracy increased from 57.37% to 63.54%, a substantial gain of 6.17%. Notably, Subject 2 exhibited the most significant improvement, with accuracy rising from 56.25% to 67.5%, an 11.25% increase. The lowest improvement was observed for Subject 5, with an increase of only 2.66%. The consistent improvements across all subjects and modifications led to statistically significant improvements ($p < 0.05$, Wilcoxon signed-rank test), emphasizing the effectiveness of the proposed approach in enhancing CS-MI classification.

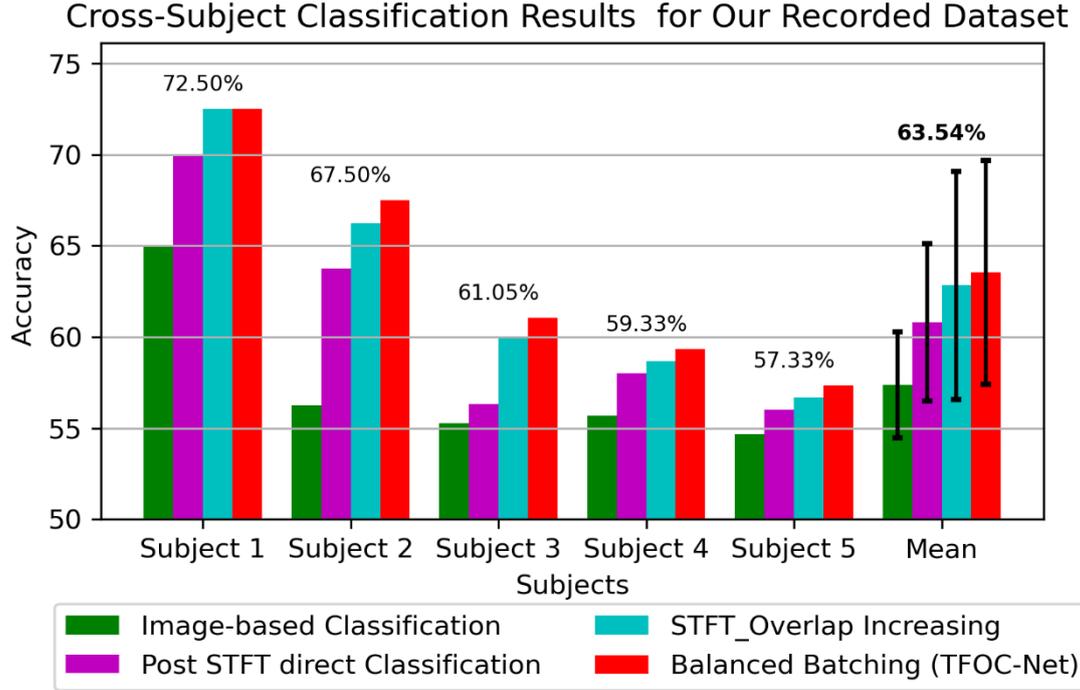

*Figure 6 Classification accuracy results on the newly recorded EEG dataset, comparing the performance of the initial image-based method with our proposed direct STFT classification approach, increased overlap, and balanced batching. The table illustrates the substantial improvements in cross-subject classification accuracy achieved with our refined methodology.*

*3.6. Comparison with the State-of-the-art Methods*

To thoroughly evaluate our proposed approach, we contrasted its classification outcomes across all three benchmark datasets with the leading results in the field of CS-MI EEG classification. **Table 5** provides a comparison of our proposed approach with several notable studies, highlighting the mean accuracies achieved and the statistical significance (p of Wilcoxon signed-rank test) for each dataset.

On dataset IV-2B, TFOC-Net reached a mean accuracy of 80.22%, notably outperforming all other listed methods. The improvement in our prior work is statistically significant ($p = 0.003$), underscoring the effectiveness of the refinements reported here. Similarly, for dataset IV-2A, we attained a mean accuracy of 65.96%. While this result did not significantly surpass our previous framework ($p = 0.492$), it did outperform other state-of-the-art techniques such as those presented in [8] and [37], with p-values less than 0.05. However, Liang et al. [38] approach achieved a slightly higher mean accuracy of 67.42%, though the difference was not statistically significant ($p = 0.734375$). It is important to note that while Liang et al. method outperformed ours in terms of overall mean accuracy, our approach demonstrated superior performance in 55.5% of the subjects (subjects 1, 2, 3, 4, and 6).

This suggests that our method might be more effective for certain individuals or groups, highlighting the variability in performance across different subject populations. It should be noted though that Liang et al. classification model is more complex than ours and the reported CS results for only one dataset. Lastly, on dataset IV-1, our approach yielded a mean accuracy of 67.60%, reaching a significance enhancement ($p = 0.062$) compared to the WGAN-CNN model. These comparative findings emphasize the strength of our proposed framework in cross-subject motor imagery EEG classification. The consistent gains across various benchmark datasets highlight the efficacy of the modifications we have made, such as direct classification after STFT, increased STFT overlap, and balanced batching. Moreover, the statistical significance of these improvements reinforces the robustness and generalizability of our approach.

*Table 5 Comparative analysis of CS classification accuracy on benchmark datasets (IV-2B, IV-2A, IV-1) against state-of-the-art methods. The table provides mean accuracies and statistical significance, highlighting the superior performance of our proposed framework in motor imagery EEG classification.*

| Study | Mean Accuracy | p-value |
|---|---|---|
| **Dataset IV-2B** | | |
| Zhu et al. [7] | 64% | N/A |
| Roy et al. [10] | 67.78% | 0.0019531 |
| Dolzhikova et al. [6] | 66.28% | 0.0019531 |
| An et al. [12] | 67.93% | N/A |
| WGAN-CNN [18] | 70.55 % | 0.003 |
| TFOC-Net | **80.22 ±0.11%** | |
| **Dataset IV-2A** | | |
| Riyad et al. [8] | < 54% | 0.0019531 |
| Amin [37] | 55.34% | 0.0058593 |
| Dolzhikova et al. [6] | 64.16% | 0.0488 |
| Lui [29] | 63.34 % | 0.431 |
| WGAN-CNN [18] | 65.27 % | 0.492 |
| Liang et al.[38] | **67.42%** | 0.734375 |
| TFOC-Net | 65.96 ± 6.14% | |
| **Dataset IV-1** | | |
| WGAN-CNN [18] | 63.7% | 0.062 |
| TFOC-Net | **67.60 ± 7.89%** | |

## 4. Discussion

The primary objective of this study was to improve the classification performance of EEG signals for CS-MI tasks through optimized preprocessing techniques and advanced classification methods using our designed TFOC-Net. We investigated the potential benefits of classifying EEG data directly using STFT compared to image-based classification. Additionally, we explored the impact of STFT parameters on the performance of MI classification from EEG signals.

Our results indicate that this direct classification using TFOC-Net can further enhance classification performance compared to image-based classification, likely due to the preservation of the full information content of the STFT results. We avoid the potential loss of information that can occur during image conversion and allow the classifier to access the raw STFT features directly. Additionally, it avoids the computations overhead as the STFT results can be passed directly to the classifier without the need for file I/O operations, particularly for Dataset IV-2B, which achieved an 11% increase in performance compared to 1.09~3.44% for the other Datasets. This significant improvement for Dataset IV-2B can be attributed to the larger number of trials available (around 720 per subject), which allowed the model to generalize better and capture finer temporal and frequency details. The richness of Dataset IV-2B data enabled CNN to fully exploit the time-frequency information in the raw STFT data, whereas converting the STFT into images likely resulted in the loss of important details.

Moreover, we found that increasing the overlap between consecutive STFT windows in TFOC-Net from minimal (1 sample) to 50% led to a significant improvement in classification accuracy across all used benchmark datasets (IV-1, IV-2A, and IV-2B). This improvement was particularly pronounced in dataset IV-1, where the mean accuracy increased from 60.29% to 67%. Increasing the overlap enhances the continuity between segments, leading to a more detailed and smoother representation of the temporal dynamics of the EEG signals. This adjustment significantly impacted on the feature extraction process and, consequently, the classification accuracy. Higher overlap results in a denser time-frequency representation, which captures transient events and subtle changes in the EEG signals more effectively [39]. For EEG data, which often contains intricate and rapid fluctuations, such detailed capture is crucial. Our findings align with existing literature that emphasizes the importance of parameter tuning in time-frequency analysis for EEG signal processing [19], [21]. The minimal overlap setting, while providing high time resolution, may have led to overfitting due to the model learning noise or irrelevant details in the training data. The 50% overlap setting appears to strike a better balance between time and frequency resolution, acting as a form of regularization and preventing the model from fitting too closely to the training data. However, 50% overlap increases computation significantly compared to minimal overlap.

In addition to STFT parameter optimization, we implemented a balanced batching strategy during training TFOC-Net to ensure that each training batch contains data from all training subjects, promoting better generalization. This strategy consistently yielded the best results across all datasets, with statistically significant improvements observed in datasets IV-2A and IV-2B. While the improvement in dataset IV-1 was not statistically significant, it still led to the highest mean accuracy.

It is important to note that our CS evaluation strictly adhered to a calibration-free paradigm, ensuring that no data from the test subjects was utilized during any stage of model training. To maintain the rigor of our CS evaluation, we deliberately omitted studies such as [40], [13], [35], and [41] from our comparative analysis. These studies, while valuable, employed methodologies that incorporated some degree of subject-specific information during the training process, even if indirectly. This inclusion, however subtle, can potentially confound the assessment of true cross-subject generalizability.

Along with these methodological advancements, we also investigated the effect of different MI time segments on classification performance. By analyzing the classification accuracy for the full 4-second window, 3-second window, as well as various 2-second and 1-second segments, we observed that the first 3 seconds of the MI task generally yielded comparable accuracy across datasets, and the middle 2 seconds performing particularly well in two datasets. This suggests that crucial brain activity for MI classification is captured early on, while the middle of the task may also provide stable, discriminative features. These findings highlight the importance of temporal dynamics in EEG signals and suggest that future approaches could optimize classification by focusing on specific time segments, potentially reducing trial length while maintaining high accuracy.

## 5. Conclusion

This study is, to our knowledge, the first to present a CS-MI classification model validated across four datasets, three of which are standard benchmarks, making the proposed model the most extensively evaluated model in this domain. This paper introduces TFOC-Net as a novel approach, utilizing EEG spectrum features and deep learning to enhance accuracy. We explored several critical factors, such as direct classification after STFT versus image-based classification, increased STFT overlap, and balanced batching during CNN training, to improve EEG signal representation and analysis, resulting in notable classification performance and generalizability. Our method, evaluated on three benchmark datasets (IV-1, IV-2A, and IV-2B) and a newly recorded EEG dataset, outperformed other state-of-the-art techniques, achieving the best-reported accuracy results. These results emphasize the potential in advancing the field of calibration-free MI-based BCIs. These findings underscore the value of optimized preprocessing and classification techniques for calibration-free MI-based BCIs. Future work will focus on exploring additional preprocessing techniques, such as artifact removal and feature selection, to further enhance the quality of EEG signals. We will also investigate the application of our approach to other BCI paradigms and real-world scenarios, aiming to develop truly user-friendly and adaptive BCI systems. By pushing the boundaries of cross-subject MI classification, we strive to unlock the full potential of this transformative technology and pave the way for a new era of brain-computer interaction.

**CRediT authorship contribution statement**

**Ahmed G. Habashi:** Methodology, Software, Validation, Formal analysis, Writing – Original Draft, Visualization. **Ahmed M. Azab**: Conceptualization, Methodology, Writing – Review & Editing, Supervision. **Seif Eldawlatly**: Conceptualization, Methodology, Writing – Review & Editing, Supervision, Project administration. **Gamal M. Aly**: Conceptualization, Writing – Review & Editing, Supervision.

**Declaration of Competing Interest**

All authors declare that they have no known competing financial interests or personal relationships that could have appeared to influence the work reported in this paper.